# A Scale and Rotational Invariant Key-point Detector based on Sparse Coding

THANH PHUOC HONG and LING GUAN, Electrical and Computer Engineering Department, Ryerson University, Canada

___

Abstract: Most popular hand-crafted key-point detectors such as Harris corner, SIFT, SURF aim to detect corners, blobs, junctions or other human defined structures in images. Though being robust with some geometric transformations, unintended scenarios or non-uniform lighting variations could significantly degrade their performance. Hence, a new detector that is flexible with context change and simultaneously robust with both geometric and non-uniform illumination variations is very desirable. In this paper, we propose a solution to this challenging problem by incorporating Scale and Rotation Invariant design (named SRI-SCK) into a recently developed Sparse Coding based Key-point detector (SCK). The SCK detector is flexible in different scenarios and fully invariant to affine intensity change, yet it is not designed to handle images with drastic scale and rotation changes. In SRI-SCK, the scale invariance is implemented with an image pyramid technique while the rotation invariance is realized by combining multiple rotated versions of the dictionary used in the sparse coding step of SCK. Techniques for calculation of key-points' characteristic scales and their sub-pixel accuracy positions are also proposed. Experimental results on three public datasets demonstrate that significantly high repeatability and matching score are achieved.

___

## 1. Introduction

Inspired by the success of corner detection works in images by Förstner [26], Harris [8], many hand-crafted key-point detectors such as SIFT [16], SURF [3], MSER [17], SFOP [7] have been proposed. These detectors have been successfully applied to a large number of vision based tasks such as image matching, image registration and object recognition. However, the aforementioned detectors are known to be inflexible to change in contexts [25, 28], likely due to the fact that the hand-crafted detectors rely on a human designed structure such as a corner, blob, or junction. In fact, a human designed structure may help make these detectors robust to some extent with some geometric transformations. However, the effectiveness of the detectors is largely limited to particular scenarios. In other words, when being used in a situation there is no sufficient high-quality key-points satisfying the pre-defined structures, the performance of these detectors will decrease. In addition to this well-known inflexibility, the robustness of the detectors has also been proved to significantly degrade under non-uniform illumination change [9]. Thus, a detector that is not only robust against geometric change, but also flexible with different contexts and invariant with non-uniform illumination variations is very desirable. Though having been very demanding, a solution to such a challenging problem has not yet been made available to our knowledge.

Recently, a Sparse Coding based Key-point detector (SCK) which relies on no pre-designed structures was proposed [9, 10]. As being independent of any human designed structure, SCK is very flexible to context changes. In SCK, a key-point is detected based on the number of non-zero components (complexity measure) in the sparse representation of the block surrounding its position. If complexity measure of a block falls in a predefined range, the center of the block is considered a good key-point. A strength measure is then utilized to compare and select key-points in the event the maximum number of key-points is limited. With the use of zero mean, unit amplitude normalization of the input blocks to the sparse coding algorithm (an intrinsic step in the sparse coding), the detector has been proved to be invariant with non-uniform illumination variations formulated by an affine intensity change model. Although SCK has demonstrated impressive performance compared with the other detectors, currently it is not designed to handle images with significant changes in scale and rotation.

In this paper, we propose to tackle the aforementioned challenges in key-point detection by incorporating scale and rotation invariant design into SCK while maintaining its original properties: flexibility and non-uniform lighting change invariance. It is worth notetaking that available scale invariant frameworks such as DoG, Harris Laplace are not used in this work as they are difficult to adapt to structures other than those they are designed upon and have been shown unable to handle non-uniform illumination variance [9]. The newly proposed detector in this work is called Scale and Rotation Invariant Sparse Coding based Key-point Detector or SRI-SCK. In SRI-SCK, the comparison of strength measure is performed at each level and then across the pyramid levels to select the most reliable key-points.

Fundamentally, the rotational invariant property is created by utilizing in the sparse coding step an extended dictionary which originates from the combination of an original dictionary and multiple rotated versions of the original dictionary atoms. In SRI-SCK, appropriate sizes/scales of key-points are automatically selected based on the comparison results of the strength measure and the levels where key-points are detected rather than being fixed as in the original SCK. We also propose to increase SCK's performance by sub-pixel accuracy estimation of SCK key-points' positions, which is based on fitting the key-points and their nearby points' strength measures to a parabola model. Experimental results on Webcam [25], VGG [19], and EF [30] datasets show that the proposed SRI-SCK detector can achieve significantly high repeatability and matching score developed by Mikolajczyk et al. [20]. To sum up, the contributions of this work are:

1. We propose a novel SRI-SCK detector that is robust with simultaneous effects of scale, rotation and non-uniform illumination change in addition to the flexibility property of SCK to different scenarios.
2. We propose an automatic means to select characteristic sizes/scales of SCK key-points.
3. We propose applying estimation of sub-pixel accuracy of key-point positions to increase SCK performance.
4. We provide a theoretical justification of various properties of SRI-SCK.

The rest of the paper is organized as follows. In Section 2, we review the algorithm of the original SCK which is called single scale SCK (SS-SCK) detector in this paper. In Section 3, SRI-SCK is presented with detailed explanation. A theoretical justification of various properties of SRI-SCK is provided in Section 4. In Section 5, the experimental results of the proposed detector are reported. The conclusion and potential future work are presented in Section 6.

## 2. A Review on Sparse Coding based Key-point Detector – SCK

The algorithm for detection of key-points in a gray-scale image using SS-SCK [9, 10] is briefly reviewed below with justification for SS-SCK's adaptability to different contexts provided in [9].

### 2.1. Sparse coding based key-point detection

After filtering the input image with a low pass filter to remove noise, each block $X$ of size $n \times n$ in the image is reshaped into a vector $X'$ of size $n^2 \times 1$, then the zero mean, unit amplitude normalization version $X'_{norm}$ of the block is calculated. Next, a sparse representation of the corresponding block is found by solving the following problem:

$$\boldsymbol{\alpha} = \text{argmin} \left( \frac{1}{2} \|X'_{norm} - D\boldsymbol{\alpha}\|_{l2}^2 + \lambda \|\boldsymbol{\alpha}\|_{l1} \right) \quad (1)$$

where $\boldsymbol{\alpha}$ is the sparse representation, $D$ is a dictionary, and $\lambda$ is the regularization parameter. The size of the dictionary is $n^2 \times k$ where $k$ is the number of atoms in the dictionary. The size of $\boldsymbol{\alpha}$ is $k \times 1$.

The number of non-zero components in the sparse representation is then defined as the complexity measure ($CM$) of the block.

$$CM = \|\boldsymbol{\alpha}\|_{l0} \quad (2)$$

If $CM$ of the block falls in a specified range, SS-SCK considers the center of this block as a key-point. The size $r$ of the key-point is calculated by (3). It is selected such that it can cover all possible positions of the key-point belonging to the block.

$$r = \left(\frac{n}{2}\right)\sqrt{2} \quad (3)$$

### 2.2. Strength measure of a key-point

$CM$ could not be used to sort and select points as many of them have the same values. Thus, a strength measure ($SM$) which is the multiplication of norm-0 and norm-1 of the sparse representation is proposed in [9] for comparison and selection of key-points.

$$SM = \|\boldsymbol{\alpha}\|_{l0} * \|\boldsymbol{\alpha}\|_{l1} \quad (4)$$

The strength measure is, however, not necessarily restricted to (4). As long as a measure satisfies certain conditions specified in [9], it is likely to be a quality $SM$.

### 2.3. Non-max suppression
In this step, only key-points whose $SM$s are the highest in their neighborhood are preserved for succeeding processes such as descriptor calculation and matching.

### 2.4. Scale selection for descriptor calculation
Scales where we locate key-points are necessary for the evaluation of SS-SCK with matching score metric [20]. Inspired by an observation that a black disk with radius $r$ in a white background gives the highest response to a normalized Laplacian of Gaussian function (5) at the scale $\sigma$ equaling to $r/\sqrt{2}$, in [9], the scale for a SS-SCK key-point is selected as $\sigma = r/\sqrt{2}$ where $r$ is the size of the key-point. This is because at each location, a key-point is estimated as a disk with radius $r$, any points inside $r$ are regarded as potential key-point positions while any points outside $r$ are not.

$$LoG = \frac{(x^2+y^2-2\sigma^2)}{\sigma^4} e^{-(x^2+y^2)/2\sigma^2} \tag{5}$$

## 3. Scale and Rotational Invariant Key-point Detector based on Sparse coding – SRI-SCK

In this section, the proposed algorithm for detection of SRI-SCK key-points in an input image is presented.

### 3.1 Construction of image pyramid and pre-processing
An image pyramid of the input image is constructed with a given scale factor. Figure 1 shows an example pyramid of one image in Webcam dataset. The input image is arranged at the bottom of the pyramid (Depending on application, a magnified version of the input image could be set at the bottom level). Each higher level in the pyramid is an image whose first and second dimensions equal to the scale factor multiplied by the corresponding dimensions of its previous level. For each level in the image pyramid, we perform low-pass filtering as in the SS-SCK to remove noise. If the image is a color one, the grayscale conversion step could be performed on the input image or at each pyramid level (before or after low-pass filtering). After grayscale conversion and low-pass filtering, each pyramid level is then converted to a floating-point image for the sparse coding step.

### 3.2 Calculation of complexity measure for every position at each pyramid level
At each level of the image pyramid, the calculation of $CM$ for a position is essentially similar to what is done in equations (1) and (2) of the algorithm presented in Section 2. However, the following three adjustments are necessary to be performed on (1) before $CM$ is calculated to make SCK rotational invariant.

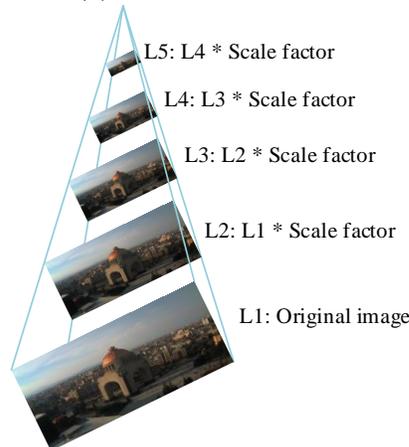

Figure 1. An example of an image pyramid

Firstly, we have to find $Y$ as a circular masked version of the block $X$. Then $X'_{norm}$ is substituted by $Y'_{norm}$ in (1). $Y'_{norm}$ is a zero mean, unit amplitude normalized version of $Y$ after being reshaped to a vector $Y'$ (elements outside the circular mask are removed before the normalization). Note: Circular masking of a block is performed by changing to zero the elements of the block that are outside a circle whose center is the center of the block.

Secondly, given that an original dictionary $D$ is selected to be used with SS-SCK, then each atom of the original dictionary can be reshaped into a block of $n$ x $n$. After being reshaped, each atom is circularly masked as what has been done with the block $X$. Each circular masked atom is then rotated to angles $\beta = \frac{360}{v}$ ; $2\beta = \frac{2*360}{v}$; ...; $(v-1)\beta = \frac{(v-1)*360}{v}$ ($v$ is a non-negative integer number) to create $(v-1)$ rotated versions of the atom. All original atoms and their rotated versions are then reshaped to vectors (elements outside the circular mask are removed) and eventually combined to create an extended dictionary $ED$. If original atoms of the dictionary $D$ are already rotationally symmetric with step $\beta$ (their circular masked versions become themselves when being rotated by an angle $\beta$), the rotated versions of the atoms are not necessary for the creation of $ED$.

Thirdly, a norm-2 penalty is added to the optimization equation. In fact, by doing the first and second adjustments, it can be shown that when $X$ is rotated by an angle $\delta = u\beta, u \in Z$ (or integer number), a solution of the revised equation is a new sparse representation $\alpha'$ in which the number and values of non-zero components are the same as $\alpha$, though the non-zero positions are shifted (see proof in Section 4.2). With this new sparse representation, the number and values of non-zero components are un-changed, so the $CM$ and $SM$ of the corresponding key-point are unaffected. This means that the detector is rotational invariant. However, with only norm-1 in (1), the optimization solver may find another solution rather than $\alpha'$ for the rotated block. By adding norm-2, the cost function becomes strictly convex [31], which means $\alpha'$ is the unique solution, and the rotational invariance is more concretely guaranteed.

Based on the aforementioned three adjustments, (1) is replaced by (6) as follows. Many algorithms such as LARs [5] can be applied to solve this optimization.

$$\boldsymbol{\alpha} = \operatorname{argmin}\left(\frac{1}{2}\|Y'_{norm} - ED\boldsymbol{\alpha}\|_{l2}^2 + \lambda_1\|\boldsymbol{\alpha}\|_{l1} + \frac{\lambda_2}{2}\|\boldsymbol{\alpha}\|_{l2}^2\right) \qquad (6)$$

The first and second adjustments are known techniques for creating rotational invariant sparse coding based features or descriptors [11, 14, 22], yet to the best of our knowledge, they have not been used in the key-point detection context and especially the stability of the number of non-zero components and their values under rotation of the blocks are not utilized. With the third adjustment, the revised equation is even more appropriate for key-point detection, as we prefer more stable $CM$ and $SM$ under the rotation of blocks, rather than just low reconstruction error as in traditional sparse coding.

Note: In situations when only scale invariance is expected and the need for significant in-plane rotational invariance can be alleviated, the abovementioned adjustments can be skipped and the $CM$ for each position could be calculated purely with (1) and (2).

### 3.3 Calculation of strength measure for every key-point at each pyramid level
After key-points are detected at each level by filtering $CM$ with a lower and upper limit as in the SS-SCK algorithm, a $SM$ is calculated for each detected key-point. In this paper, $SM$ (5) introduced in [9] is adopted.

### 3.4 Non-maxima suppression and estimation of key-point positions in sub-pixel accuracy for each pyramid level
After performing non-maxima suppression, we have the positions of the locally strongest SRI-SCK key-points in each pyramid level in pixel accuracy (A low-pass filter could be operated on the $SM$ image at each level to remove unstable peaks before applying non-max suppression, potentially improving the performance of SS-SCK as well as SRI-SCK). To further improve the performance of the SRI-SCK detector, we propose applying sub-pixel accuracy estimation [4] of the key-point (or the local maxima) positions in each level. The estimation process is done separately for row and column coordinates of key-points. For each key-point row coordinate, we fit the $SM$s of the corresponding key-point and its nearby points to a parabola model as in (7).

$$SM = ax^2 + bx + c \tag{7}$$

where $x$ is the row coordinate in the local Cartesian coordinate systems of the key-point, in which $x = 0$ is at the center of the pixel accuracy key-point position. Let the $SM$s for pixel at $x = 0$, pixels on the left $x = -1$ and right $x = +1$ of the key-point be $SM_{-1}, SM_0, SM_{+1}$. Replacing $x = -1, 0, 1$ into (7), we have the following three equations.

$$SM_{-1} = a - b + c \tag{8}$$
$$SM_0 = c \tag{9}$$
$$SM_{+1} = a + b + c \tag{10}$$

From (8) to (10), we can calculate $a, b, c$ of the model. The row coordinate of the maxima or the key-point in sub-pixel accuracy is then located at where the derivative of the parabola equals to zero. This is shown in the following equation.

$$x = -\frac{b}{2a} = \frac{SM_{+1} - SM_{-1}}{4SM_0 - 2(SM_{+1} + SM_{-1})} \tag{11}$$

The sub-pixel accuracy estimation process is then repeated similarly for the column coordinate of the key-point. After repeating the process for all key-points, we convert the estimated sub-pixel accurate positions in the current level to the coordinates in the first level global coordinate system.

**3.5 Sizes/scales of key-points detected at level l of the pyramid image**

The sizes of key-points detected at level $l$ of the pyramid image are calculated using the following formula.

$$s_l = \frac{s_1}{(sf)^{l-1}} \tag{12}$$

where $s_l$ is the size of a key-point detected at level $l$, and $sf$ is the scale factor for the construction of the image pyramid. This size is used for calculation of repeatability metric. If following SS-SCK, $s_1$ should be calculated using (3). However, this will make the sizes of key-points in higher levels become unexpectedly large, potentially resulting in random repetition of a specific key-point by another. To account for this situation, we use a smaller $s_1$ as $\frac{\sqrt{2}}{4}$ of the first dimension of the block size in our experiments.

Regarding the scales of key-points for descriptor calculation when evaluating our detector with matching score [20], we follow [9] to select the scale for a SRI-SCK key-point as $\sigma_l = s_l/\sqrt{2}$ where $s_l$ is the size of the corresponding key-point calculated by (12).

**3.6 Automatic selection of characteristic sizes/scales for key-points across pyramid levels of the input image**

After converting all key-point positions to coordinates of the first level in the image pyramid, a key-point A with scale $\sigma_l$ is suppressed if the $SM$ of A at level $l$ is not the highest one compared with any key-points which share a sufficiently large percentage of overlapped area with A (including A with different sizes/scales). This step further removes not only less distinctive key-point positions, but also non-representative sizes/scales of the key-points across the image pyramid. After this step, any survived positions are considered the most reliable key-points for succeeding steps like descriptor calculation or matching. Also, any survived scales/sizes associated with the key-points are selected as their characteristic scales/sizes. This scale/size selection technique is, in fact, similarly used in object detection [6].

Note: As nearby pixels in higher levels of the image pyramid are interpolated from highly overlapped neighborhoods, the blocks in higher levels tend to be flat, and the $SMs$ of key-points in these levels are expected to be smaller than those in lower levels. Thus, selection of key-points and their associated characteristic sizes/scales across pyramid levels based on comparing the $SM$ may make the detector favor key-points at lower levels. A solution inspired by [16], [18] is to apply a scale normalization function H as in (13) on the $SM$ of level $l$ and perform the current step with the scale-normalized $SM$. However, in the scope of this paper we choose $H(l, sf, SM) = SM$ when conducting experiments.

$$Scale - normalized - SM = H(l, sf, SM) \tag{13}$$

# 4. Theoretical Analysis of The Proposed SRI-SCK Key-point Detector

**4.1 The ability to maintain the illumination invariant property of SRI-SCK**
As will be demonstrated later in the experiments, SRI-SCK is very robust against significant changes in illumination as its precursors [9, 10]. This is because for each block $\boldsymbol{X} = \boldsymbol{X}(x, y, l)$, the input to the sparse coding algorithm is not its original, but the normalized version $\boldsymbol{Y'}_{norm}$ of its circular masked version $\boldsymbol{Y}$ after being reshaped to a vector $\boldsymbol{Y'}$. The formula of normalization step is as below:

$$\boldsymbol{Y'}_{norm} = \frac{Y' - \mu_Y}{\|Y' - \mu_Y\|} \tag{14}$$

where $\boldsymbol{\mu_Y}$ is the mean vector of pixel intensities inside the $\boldsymbol{Y'}$.

In computer vision, the effect of illumination change in image pixels can be modeled as below (The so-called affine intensity change model) [13, 21, 27, 29]:

1. Each pixel intensity $\boldsymbol{I}$ changes to $a\boldsymbol{I} + b$ ($a$>0: multiplicative, and $b$: additional effects of light) under illumination change.
2. The light effects in a small neighborhood are uniform.

Suppose that the supporting neighborhood in the original image for interpolation of the block $\boldsymbol{X}$ satisfies the aforementioned two assumptions and nearest neighbor interpolation method is used (In practice, good results could also be generated by different interpolation method such as bi-cubic interpolation), under different lighting conditions, the block $\boldsymbol{X}$ after change could be modeled as $a\boldsymbol{X} + b$. The circular mask version $\boldsymbol{Y}$ after change then becomes $a\boldsymbol{Y} + b$. However, the input of the sparse coding step, remains unchanged (15).

$$\frac{(aY'+b)-(a\mu_Y+b)}{\|(aY'+b)-(a\mu_Y+b)\|} = \frac{Y'-\mu_Y}{\|Y'-\mu_Y\|} = Y'_{norm} \qquad (15)$$

As the input is the same, the $CM$ and $SM$ are thus unaffected under affine intensity change. <u>This confirms that SRI-SCK is able to maintain its robustness under non-uniform illumination variations as SS-SCK.</u>

### 4.2 Rotational invariant property of the SRI-SCK

To make SS-SCK rotational invariant, we need to re-design the dictionary such that $CM$ and $SM$ of a circular masked block are invariant with rotation of the block. This can be achieved by combining original circular masked atoms and their rotated versions in an original dictionary to create a new extended dictionary. This approach requires only one round of sparse coding for each point and the current structure of SS-SCK can be re-used, so it is favorable for this study. The steps for handling rotational invariance in SRI-SCK have been presented in Section 3. In this section, the property is proved as below.

First, as explained in Section 3, if the current dictionary of SS-SCK (all atoms are circular masked, elements outside the mask are removed) is $\boldsymbol{D} = [\boldsymbol{D_{01}}, \dots, \boldsymbol{D_{0k}}]$, then the dictionary for SRI-SCK is as follow

$$\boldsymbol{ED} = [\boldsymbol{D_{01}}, \dots, \boldsymbol{D_{0k}}, \dots, \boldsymbol{D_{(v-1)1}}, \dots, \boldsymbol{D_{(v-1)k}}] \qquad (16)$$

where $\boldsymbol{D_{ij}} = \mathbf{Rot}(\boldsymbol{D_{(i-1)j}}, \beta) = \mathbf{Rot}(\boldsymbol{D_{0j}}, i\beta)$. In other words, $\boldsymbol{D_{ij}}$ is the rotated version of $\boldsymbol{D_{(i-1)j}}$ by an angle $\beta = \frac{360}{v}$ or rotated version of $\boldsymbol{D_{0j}}$ by an angle $i\beta = \frac{i360}{v}$. Meanwhile, $\boldsymbol{D_{0j}}$ is the rotated version of $\boldsymbol{D_{(v-1)j}}$ by the angle $\beta$.

Now, suppose that before the rotation of a circular masked block $Y$, we have a solution $\boldsymbol{\alpha}$ for (6) as follows.

$$\boldsymbol{\alpha} = \mathrm{argmin}(\tfrac{1}{2}\|Y'_{norm} - [\boldsymbol{D_{01}}, \dots, \boldsymbol{D_{0k}}, \dots, \boldsymbol{D_{m1}}, \dots, \boldsymbol{D_{mk}}]\boldsymbol{\alpha}\|_{l2}^2 + \lambda_1 \|\boldsymbol{\alpha}\|_{l1} + \tfrac{\lambda_2}{2}\|\boldsymbol{\alpha}\|_{l2}^2) \qquad (17)$$

where $m = v - 1$. In other words, with a sparse representation $\boldsymbol{\alpha} = [\boldsymbol{\alpha_{01}}, \dots, \boldsymbol{\alpha_{0k}}, \dots, \boldsymbol{\alpha_{m1}}, \dots, \boldsymbol{\alpha_{mk}}]^T$, the cost function in the optimization problem achieves the minimum.

$$s = \tfrac{1}{2}\|Y'_{norm} - [\boldsymbol{D_{01}}, \dots, \boldsymbol{D_{0k}}, \dots, \boldsymbol{D_{m1}}, \dots, \boldsymbol{D_{mk}}]\boldsymbol{\alpha}\|_{l2}^2 + \lambda_1 \|\boldsymbol{\alpha}\|_{l1} + \tfrac{\lambda_2}{2}\|\boldsymbol{\alpha}\|_{l2}^2 = s_{min} \qquad (18)$$

For an arbitrary point in $Y'_{norm}$, we have the following intensity equation:

$$I(\rho, \beta_0) = \alpha_{01} I_{01}(\rho, \beta_0) + \cdots + \alpha_{0k} I_{0k}(\rho, \beta_0) + \cdots + \alpha_{m1} I_{m1}(\rho, \beta_0) + \cdots + \alpha_{mk} I_{mk}(\rho, \beta_0) + e \qquad (19)$$

where $(\rho, \beta_0)$ are the radial and angular coordinates of the point and the atoms when $Y'_{norm}$ and the dictionary atoms are represented in circular forms, $e$ is the reconstruction error of the point. The reconstruction error remains the same for this point, if we rotate the circular masked input block and the dictionary atoms by the same angle $u\beta$ ($u$ is an integer) (20).

$$I(\rho, \beta_0 + u\beta) = \alpha_{01} I_{01}(\rho, \beta_0 + u\beta) + \cdots + \alpha_{0k} I_{0k}(\rho, \beta_0 + u\beta) + \cdots + \alpha_{m1} I_{m1}(\rho, \beta_0 + u\beta) + \cdots + \alpha_{mk} I_{mk}(\rho, \beta_0 + u\beta) + e \qquad (20)$$

The second part of (20) can be rewritten below.

$$I(\rho, \beta_0 + u\beta) = \alpha_{01} I_{u1}(\rho, \beta_0) + \cdots + \alpha_{0k} I_{uk}(\rho, \beta_0)$$

$$+ \cdots + \alpha_{m1} I_{(m+u)1}(\rho, \beta_0) + \cdots +$$

$$\alpha_{mk} I_{(m+u)k}(\rho, \beta_0) + e$$

$$= \alpha_{01} I_{u1}(\rho, \beta_0) + \cdots + \alpha_{0k} I_{uk}(\rho, \beta_0)$$

$$+ \cdots + \alpha_{m1} I_{(v-1+u+1-1)1}(\rho, \beta_0) + \cdots +$$

$$\alpha_{mk} I_{(v-1+u+1-1)k}(\rho, \beta_0) + e$$

$$= \alpha_{01} I_{u1}(\rho, \beta_0) + \cdots + \alpha_{0k} I_{uk}(\rho, \beta_0)$$

$$+ \cdots + \alpha_{m1} I_{(u-1)1}(\rho, \beta_0)$$

$$+ \cdots + \alpha_{mk} I_{(u-1)k}(\rho, \beta_0) + e \tag{21}$$

From (21), we can see that for a block rotated by an angle $u\beta$, if we perform circular shift of the components of $\boldsymbol{\alpha}$ (found before the rotation) by $u$ units (atom wise), we have a new representation $\boldsymbol{\alpha}'$ maintaining the same reconstruction error as before the rotation of the block. As no change occurs in the number of non-zero components and their values, the cost function remains the minimum $s_{min}$. The cost function in (18) is strictly convex, so $\boldsymbol{\alpha}'$ is the unique sparse representation for the rotated block. $\boldsymbol{\alpha}'$ and $\boldsymbol{\alpha}$ have the same numbers of non-zero components which have changes in positions but not in values. Hence, the $CM$ and $SM$ of the block are unaffected. In other words, the detector is invariant with rotation of the block.

The concept of the above verification is illustrated with Figure 2, where the normalized circular masked version of the block and the atoms of the $\boldsymbol{ED}$ are plotted in their circular shapes. There are four steps in the proof. In Figure 2a, we suppose that the input block is represented by a sparse representation $\boldsymbol{\alpha}$ solved from (6). In Figure 2b, without loss of generalization, we assume the input block is rotated by an angle $1\beta$, the same sparse representation $\boldsymbol{\alpha}$ will be achieved from (6) if we solve it with the atoms of the $\boldsymbol{ED}$ rotated by the same angle $1\beta$. In Figure 2c, we identify that each atom of $\boldsymbol{ED}$ after being rotated by the angle is, in fact, another atom of $\boldsymbol{ED}$. In Figure 2d, we rearrange the atoms associated with their coefficients in the original order of $\boldsymbol{ED}$. From Figure 2d, it is clear that the coefficients are shifted from atoms to others, but

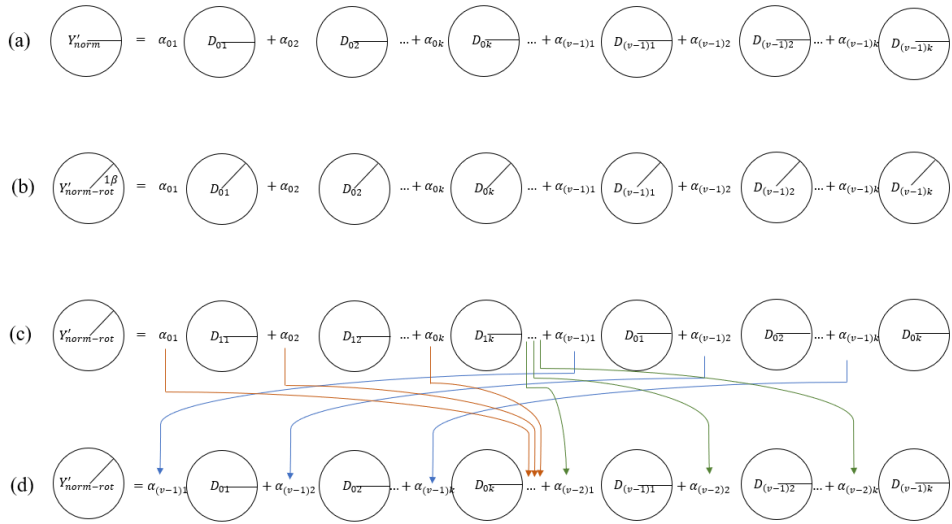

Figure 2 An illustration for the proof of rotational invariance of the key-point detector.

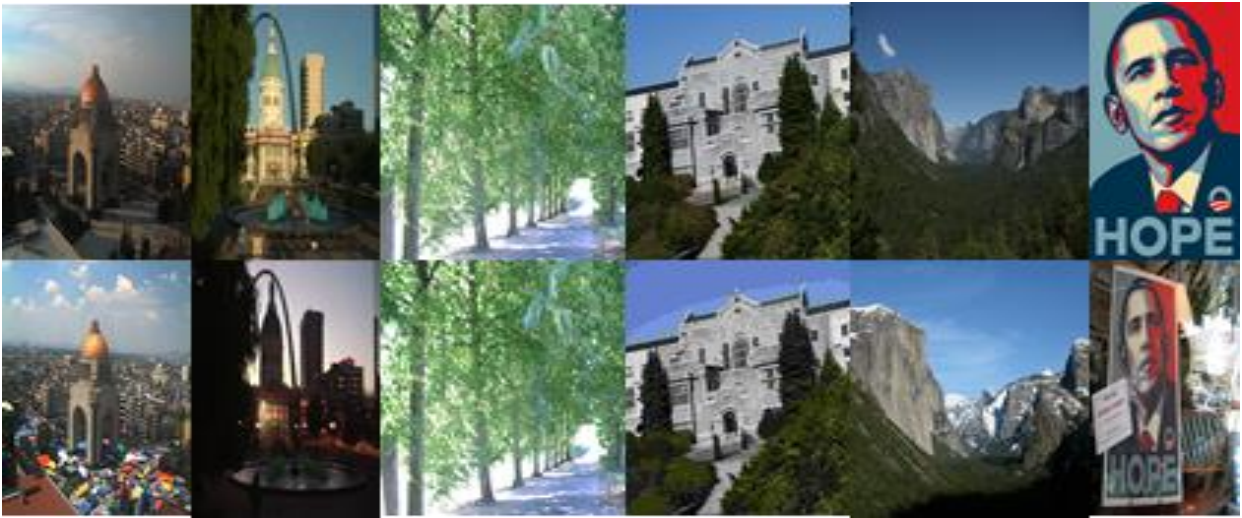
Figure 3. Examples from the three datasets (Four on the left from Webcam, four in the middle from VGG, four on the right from EF).

not changed in values, this supports the previous conclusion that $CM$ and $SM$ are not influenced, or the detector is invariant with the rotation of the block.

# 5. Experimental Results

### 5.1 Experimental datasets
The SRI-SCK detector is evaluated on three public datasets: Webcam [25], VGG [19] and EF [30] (some sample images are shown in Figure 3):

1. Webcam dataset [25]: This dataset consists of 6 sequences, each of which has 140 images of the same scene (Training: 100; validation: 20; testing: 20). In this dataset, each pair of scenes features variations in time and seasons. Only performance of testing images is reported in this paper.

2. VGG dataset [19]: This dataset consists of 8 sequences each of which has 6 images featuring different levels of change in blur, viewpoint, scale, rotation angle, lighting, and JPEG compression.

3. EF dataset [30]: This dataset consists of 5 sequences of images with wide ranges of changes in scale, illumination and background clutters.

### 5.2 Settings for SRI-SCK in experiments
The settings for evaluating SRI-SCK are shown in Table 1. Multiple settings are provided to give some perspectives about the performance of SRI-SCK with different sets of parameters. In the settings, a scale factor 0.8 is used. The "Ext-DCT-2" dictionary (Figure 4) for SRI-SCK is an extended dictionary created from a diagonal atom in a DCT-2 basis [12, 23] ((22) shows the formula of the basis for one element $(e,f)$th $(1 \leq e, f \leq n)$ in one atom $i$th $(i = (p-1)n + q, 1 \leq p, q \leq n)$ of its original size $n$x$n$).

$$D_{i(e,f)} = a_p a_q \cos\left(\frac{\pi(2(e-1)+1)(p-1)}{2n}\right) \cos\left(\frac{\pi(2(f-1)+1)(q-1)}{2n}\right) \quad (22)$$

where $a_{p,q} = 1/\sqrt{n}$ when $p, q = 1$. Otherwise, $a_{p,q} = \sqrt{2/n}$. It is observed that the circular masked version of the atom (Top left one in Figure 4) is rotationally symmetric with a 90 degree step (it becomes itself after being rotated by 90 degree), so only 8 times of rotation, with 10 degrees/each time are necessary to create the "Ext-DCT-2" that can cover 360 degree rotation. Since there are only a few atoms in "Ext-DCT-2", no upper and lower limit are set for the SRI-SCK settings. Additionally, as random matching of key-points may occur if a large number of key-points are used, maximum 1000 key-points/image are considered as in [28]. Note: as the implementations of some of the methods for comparison with SCK such as Harris Laplace do not have a clearly associated strength measure, in our previous works, we followed [28] in which average 1000 key-points/image is used for calculation of the quantitative results of these methods, causing potential inconsistency in quality measurement. Through substantial experiments, we have identified (and used in this work) that a corner-ness measure provided with the implementations in [24] is suitable for sorting and selecting the points for these methods. Besides, we studied another issue associated with some methods in [28] that generate very few key-points and introduced appropriate thresholds with proper adjustment to gain sufficient points for the selection. With the abovementioned steps, we expect to generate fairer perspectives in performance evaluation and comparison conducted in this paper.

Table 1. Configurations of SRI-SCK detector

| Name | Dictionary | Block Size | $\lambda_1, \lambda_2$ |
|---|---|---|---|
| SRI-SCK-1 | Ext-DCT-2 | 21 × 21 | 0.125, 0.375 |
| SRI-SCK-2 | Ext-DCT-2 | 25 × 25 | 0.0625, 0.1875 |

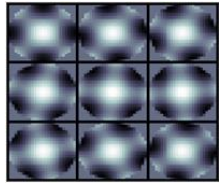

Figure 4. An extended dictionary for SRI-SCK (Images were resized for visualization)

### 5.3 Qualitative experiments

In this subsection, qualitative results of the SRI-SCK detector are presented. In this experiment, matching of key-points are performed with SIFT descriptor and Euclidean distance. In the matching scheme, two key-points $a$, $b$ in a pair of images are considered matched if the Euclidean distance between two corresponding descriptors $D_a$ and $D_b$ multiplied by a threshold (1.3) is less than the distance of $D_a$ to other descriptors of other candidate matching points [16]. The matches are then verified with ground truth transformation. Figure 5 shows three pairs of images from VGG, Webcam and EF dataset respectively featuring change in scale, rotation and/or illumination (we limit to three examples due to the space limitation). Figure 6 shows the matching results of several techniques on the VGG pair (the green and red lines respectively correspond to correct and wrong matches of key-points). As shown in Figure 6a, the matching results are very promising with SRI-SCK as the technique shows a large number of correct matches compared with others. For this pair, there is no noticeable difference between SIFT and SFOP. Harris, meanwhile, does not show good performance. Indeed, the percentage of correct matches is led by SRI-SCK (71.1%), then followed by SFOP (62.0%), SIFT (64.0%), and Harris (6.9%).

The same procedure is then performed the pair of images from Webcam dataset which feature significant variations in illumination. It is observed from Figures 7a to 7d that apparently SRI-SCK generates the best result (52.3% correct matches). Meanwhile, the number are 46.4%, 23.8%, 23.5% respectively for SFOP, SIFT, and Harris. From this pair, we can clearly see that SRI-SCK can outperform its counterparts even with significant difference of lighting conditions of imaging. The experiment is repeated on the pair of images taken from EF which have change in illumination and scale. The top percentage of correct matches (60.0%) is led by SRI-SCK, and then followed by 36.3%, 27.3%, 19.4% correct matches generated by SFOP, SIFT, and Harris respectively. The results are respectively shown in Figures 8a-8d. For the latter two experiments, as SIFT orientation assignment does not demonstrate good performance for significant illumination change, downward directions are selected for key-point orientations. To sum up, based on the qualitative experimental results, we observe that the proposed SRI-SCK is able to deliver very reliable key-points under different illumination conditions, rotations and scales. The following quantitative results will further verify this observation.

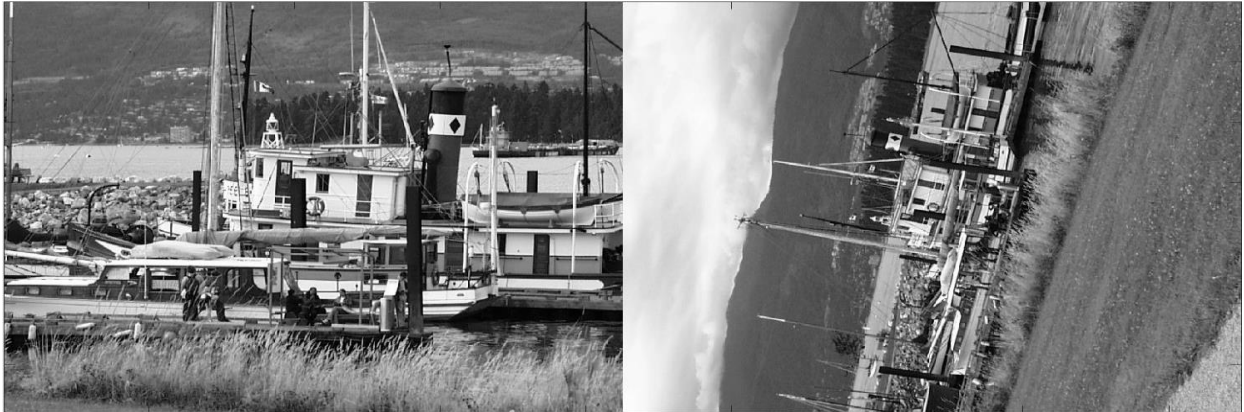
(a) A pair of images from VGG dataset

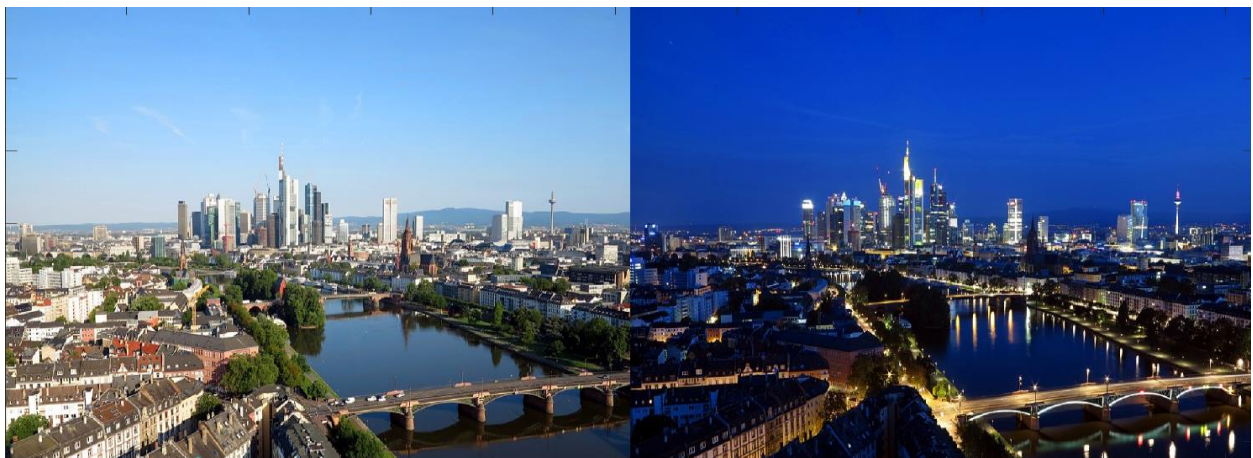
(b) A pair of images from Webcam dataset

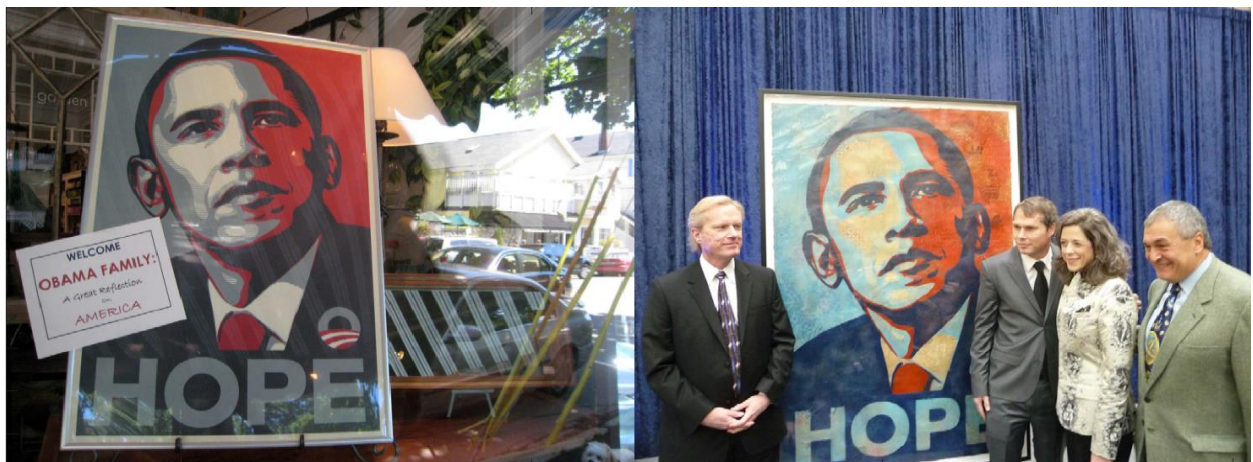
(c) A pair of images from EF dataset

Figure 5. Original pairs in qualitative experiments

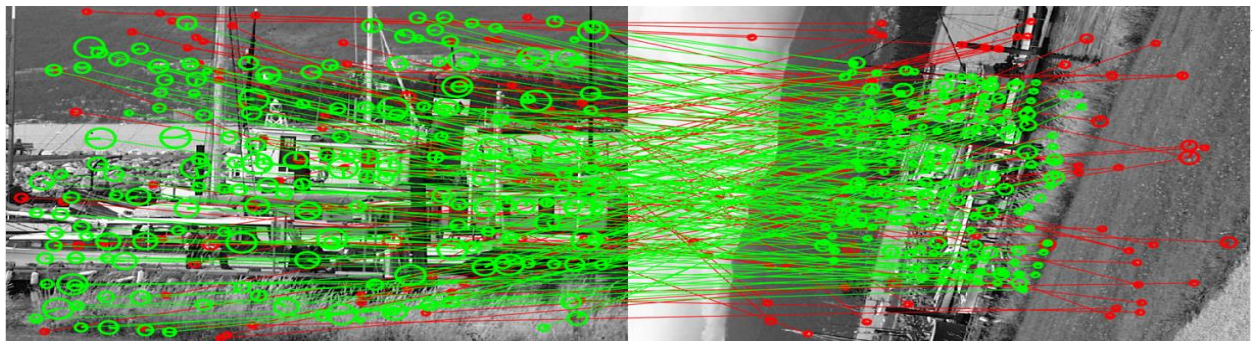
(a) SRI-SCK Detector & SIFT Descriptor - 71.1% correct

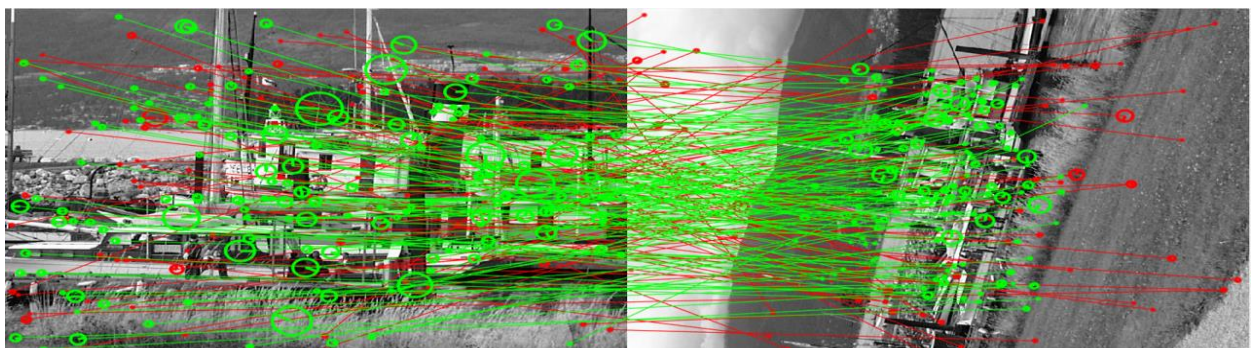
(b) SIFT Detector & SIFT Descriptor - 64.0% correct

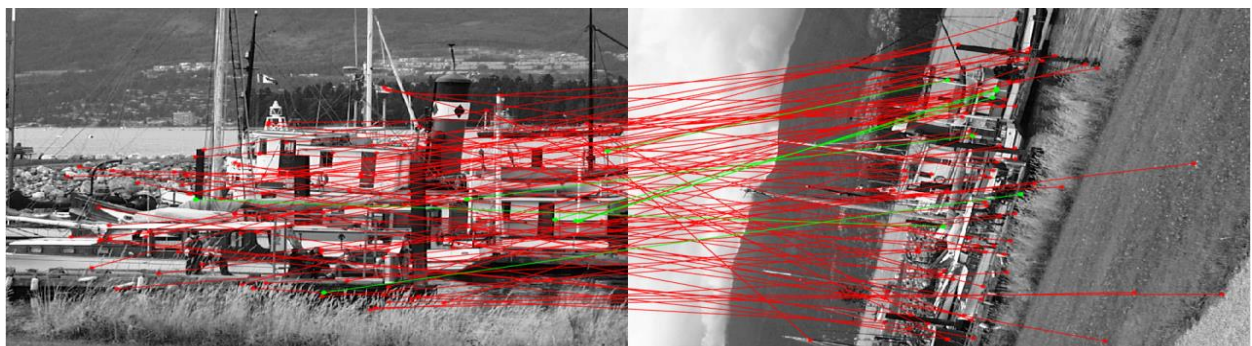
(c) Harris Detector & SIFT Descriptor - 6.9% correct

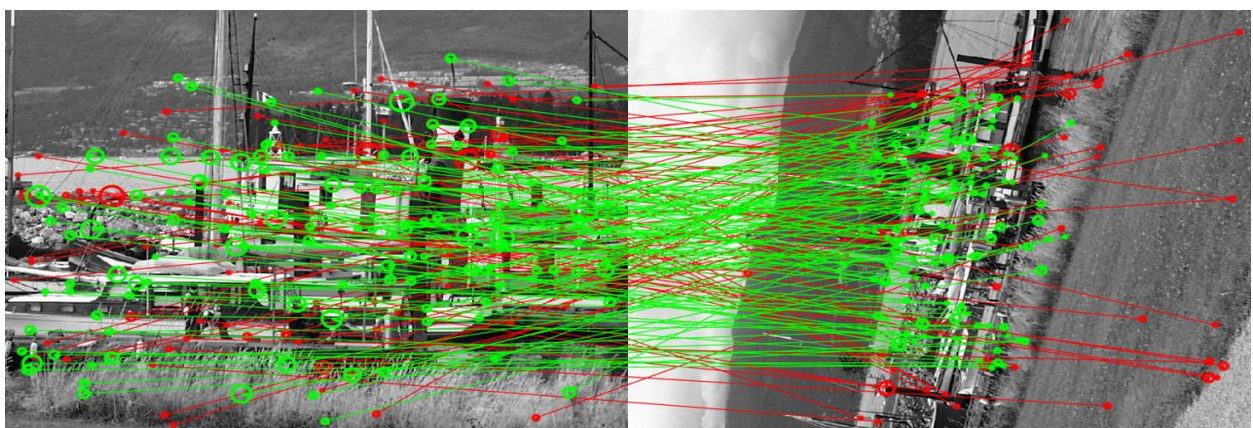
(d) SFOP Detector & SIFT Descriptor - 62.0% correct

Figure 6. An example on matching a pair of images from VGG dataset with different techniques

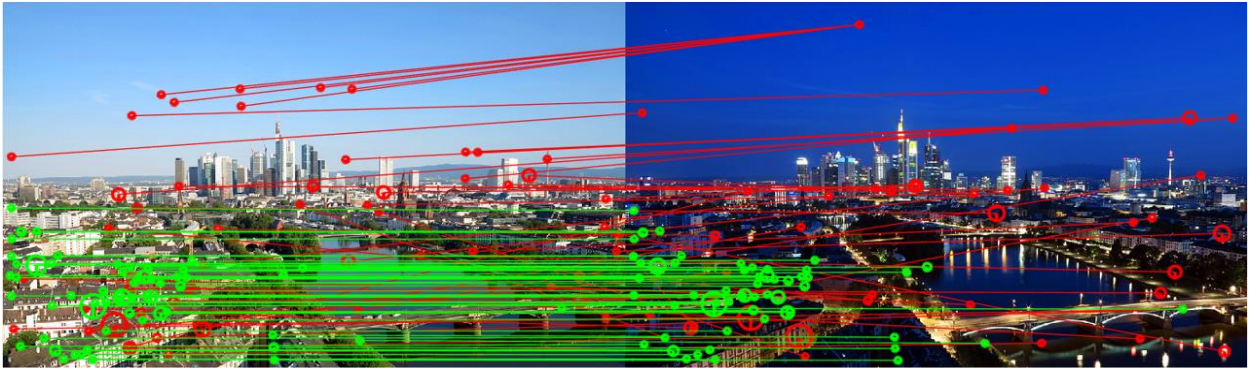
(a) SRI-SCK Detector & SIFT Descriptor – 52.3% correct

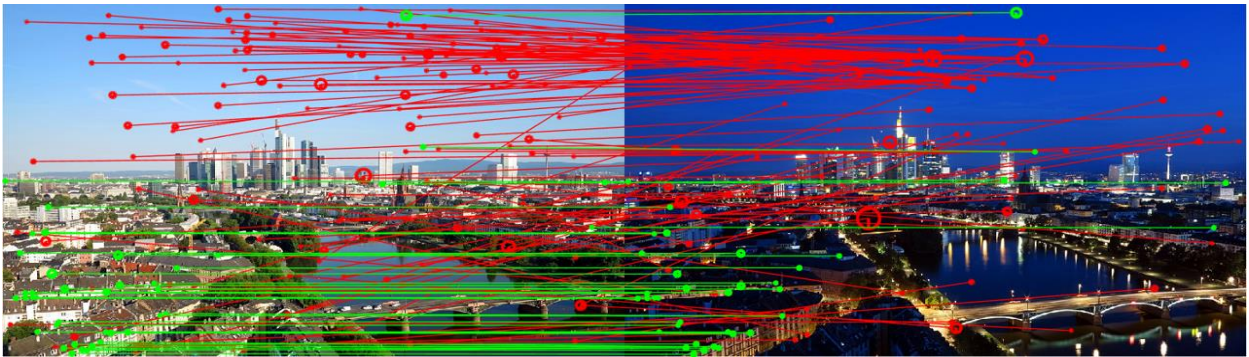
(b) SIFT Detector & SIFT Descriptor – 23.8% correct

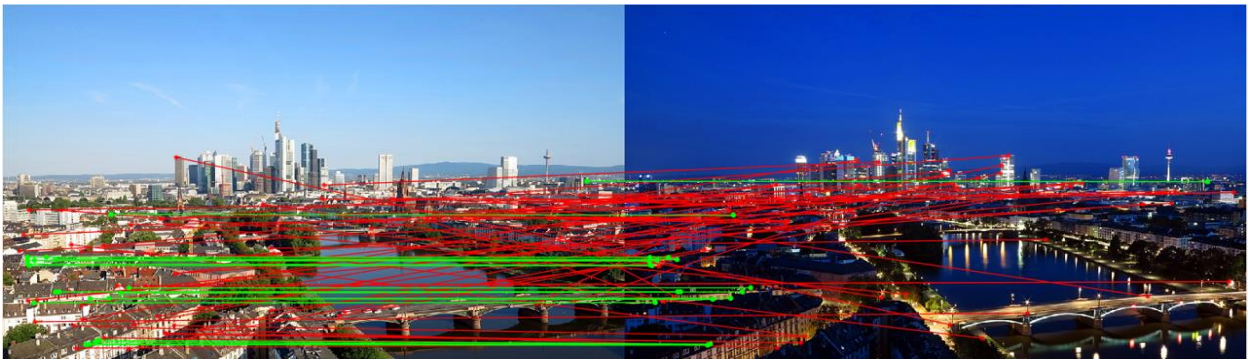
(c) Harris Detector & SIFT Descriptor – 23.5% correct

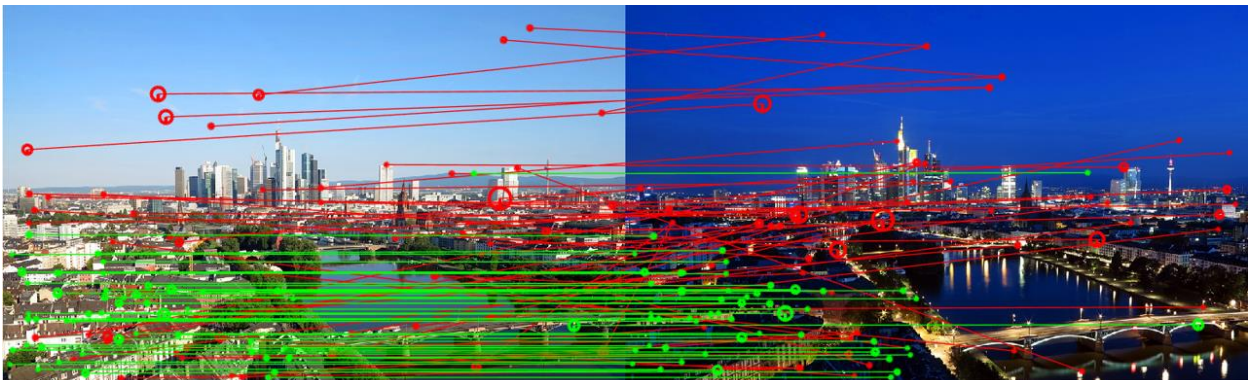
(d) SFOP Detector & SIFT Descriptor – 46.4% correct

Figure 7. An example on matching a pair of images from Webcam dataset with different techniques

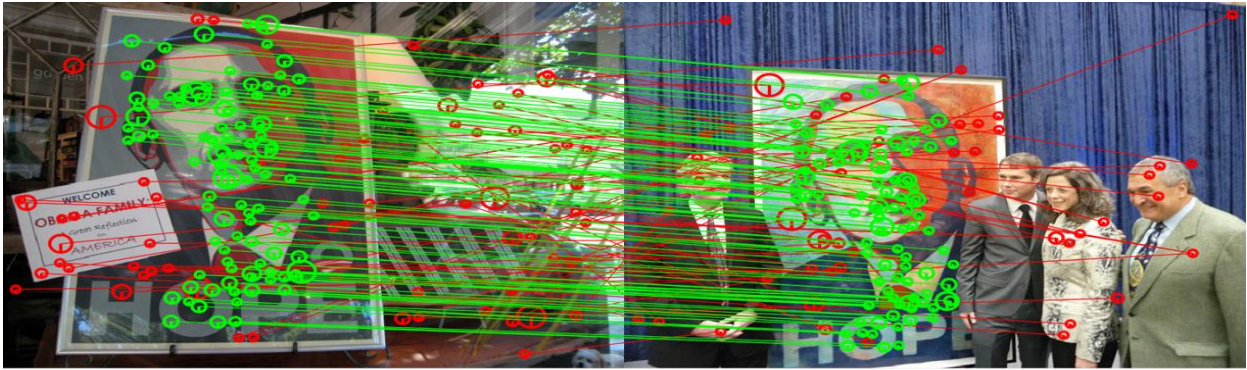
(a) SRI-SCK Detector & SIFT Descriptor – 60.0% correct

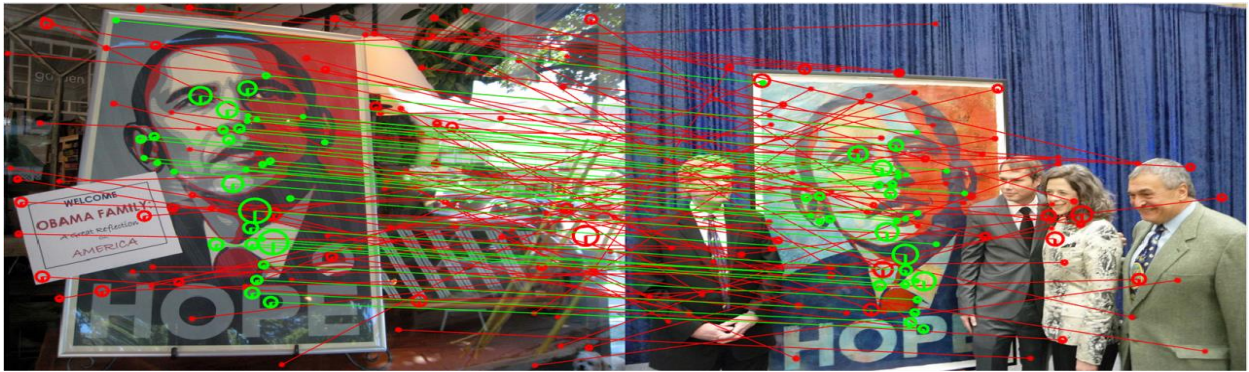
(b) SIFT Detector & SIFT Descriptor – 27.3% correct

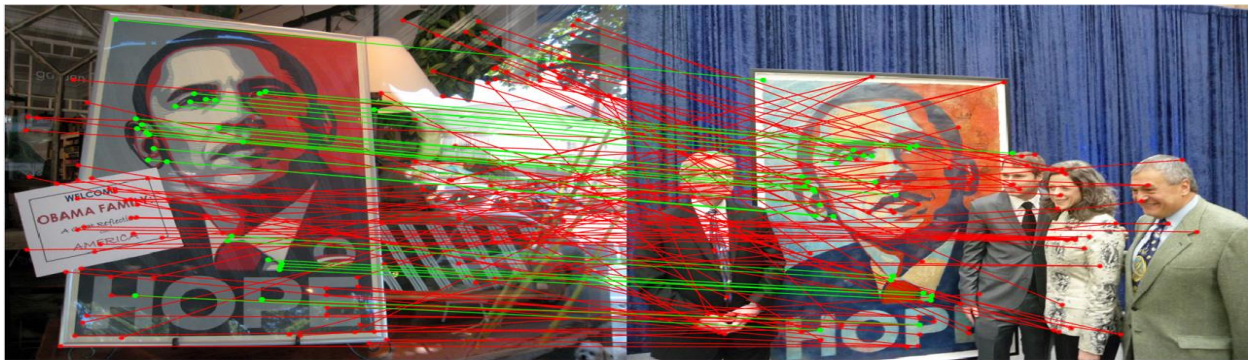
(c) Harris Detector & SIFT Descriptor – 19.4% correct

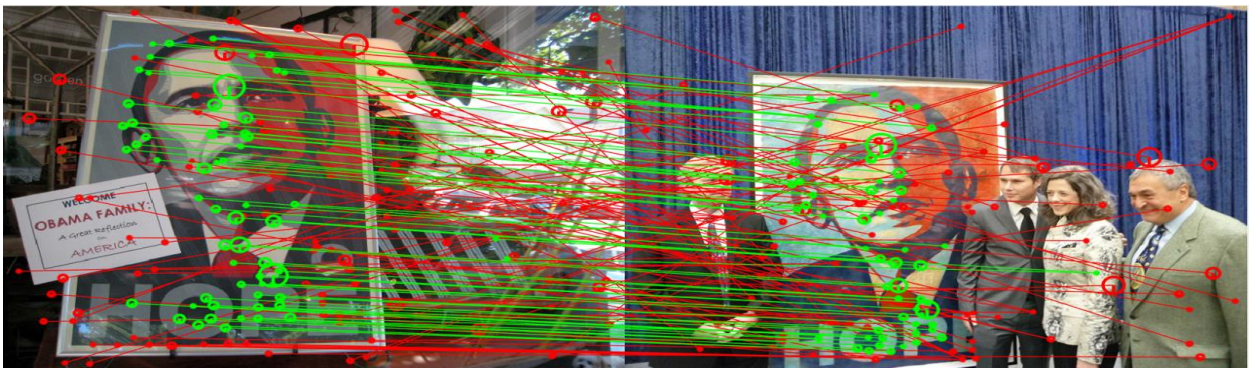
(d) SFOP Detector & SIFT Descriptor – 36.3% correct

Figure 8. An example on matching a pair of images from EF dataset with different techniques

Table 2. Repeatability of Different Methods on Various Datasets

| Type | Method | Webcam | EF | VGG |
|---|---|---|---|---|
| Hand-crafted | SIFT | 28.5 | 20.6 | 46.8 |
| | SURF | 45.5 | 40.3 | 61.1 |
| | SFOP | 38.1 | 33.6 | 49.2 |
| | KAZE | 45.5 | 37.9 | 66.4 |
| | AKAZE | 55.2 | 39.5 | 63.8 |
| | Hessian Laplace | 35.9 | 23.4 | 55.9 |
| | Harris Laplace | 38.8 | 22.2 | 52.2 |
| | Hessian Affine | 36.4 | 22.0 | 53.2 |
| | Harris Affine | 40.6 | 20.7 | 51.8 |
| | **SRI-SCK-1** | **65.1** | **47.9** | **68.1** |
| | **SRI-SCK-2** | **65.9** | **46.1** | **69.0** |
| Learning based | T-P24 | 61.1 | 46.1 | 64.2 |
| | TransCovDet | 63.4 | 49.2 | 67.3 |

Table 3. Matching score of Different Methods on Various Datasets

| Method | Webcam | EF | VGG |
|---|---|---|---|
| SIFT | 11.8 | 8.9 | 27.7 |
| SURF | 16.2 | 8.3 | 39.9 |
| KAZE | 23.7 | 12.3 | 45.7 |
| AKAZE | 26.0 | 13.7 | 46.2 |
| Hessian Laplace | 14.2 | 8.0 | 33.5 |
| T-P24 | 22.2 | 9.0 | 45.2 |
| TransCovDet | 29.2 | 11.9 | 49.8 |
| **SRI-SCK-1** | **31.0** | **14.3** | **51.0** |
| **SRI-SCK-2** | **32.3** | **14.1** | **52.3** |

**5.4 Quantitative results**

**5.4.1 Evaluation metrics:** Repeatability and matching score are used for the quantitative evaluation in this paper. As specified in [20], if the overlapped error of one key-point region in an image and the projected region from another image is less than a threshold (0.4), the two key-point regions are considered corresponding ones given that the transformation between the pair of images is known. The ratio between the number of corresponding regions (or matches) and the smaller number of regions in the shared part of the image pair is then defined as repeatability. Meanwhile, the matching score is the ratio between the number of correct matches and the smaller detected region number. A match is considered correct when the distance between two corresponding descriptors is the minimum in the descriptor space. Evaluation code in VLBenchmark [15] is used for evaluation of matching score as in [28].

**5.4.2 Other implemental details and quantitative results:** The average repeatability measures of the SRI-SCK detector on Webcam, VGG and EF datasets are compared with many popular hand-crafted detectors such as SIFT [16], SURF [3], SFOP [7], Hessian Laplace [18], Harris Laplace [18], Hessian Affine [18], Harris Affine [18], KAZE

[1], AKAZE [2] and additionally two top learning based detectors T-P24 (the best version of TILDE), and the state-of-the-art work [28] which we call Transformation Covariant Detector (TransCovDet). The performance of learning-based detectors is for reference only, as our work should be classified as a traditional detector. As illustrated in Table 2, SRI-SCK works much better than all hand-crafted detectors by a large margin (~10%, 8%, 3% higher than the top detectors: AKAZE, SURF, KAZE) respectively on the three datasets. Regarding the comparison with learning techniques, SRI-SCK also shows favored performance with TransCovDet by overcoming this detector on Webcam and VGG datasets while maintains comparable performance on EF dataset.

For the matching score, the proposed settings are compared with six representative hand-crafted detectors from the previous table including SIFT, SURF, KAZE, AKAZE, Hessian Laplace as well as the two learning based methods. Experimental studies show that SIFT orientation assignment does not work well with large non-uniform illumination change for all the methods compared in this work, the matching scores in this work are thus calculated with only downward direction of key-points for all methods (as two thirds of the testing datasets features significant non-uniform change in lighting). In [28, 15], average matching score for each $p$ image sub-dataset was calculated on first $p-1$ image pairs only, apparently generating bias and incomplete statistics. In this work, the score is calculated based on all image pairs (($p-1$) $p$ pairs) in the sub-dataset for all the methods. We are confident that, given all these adjustments made in the experiments, we expect a more justified and meaningful conclusion form the evaluation and comparison. Table 3 confirmed that SRI-SCK outperforms all the methods compared.

Note: For datasets with non-uniform lighting change, key-points at higher levels have more chance to violate the assumptions in the affine intensity change model and the cross-scale suppression helps eliminate some of unstable points among these. This is because currently the suppression is based on comparison of the $SM$ which tends to favor lower scale key-points. For solely geometric or uniform lighting change datasets such as VGG, more high scale key-points in higher pyramid levels could be useful, thus to avoid excessive suppression in these datasets, currently, a n efficient possible strategy is to scale normalize a key-point's $SM$ by multiplying it with the size/scale of the key-point. This strategy, for example, could boost the repeatability of SRI-SCK to 71.1 in VGG dataset.

## 6. Conclusions

A novel scale and rotational invariant key-point detector is proposed in this paper. The proposed detector helps widen the application range of the original SCK to work with datasets featuring simultaneous effects of drastic changes in scale and rotation in addition to non-uniform photometric transformations. The proposed detector has demonstrated remarkable performances on three benchmark datasets. Interesting future extensions of this work could be an algorithm which incorporates an affine region adaptation step [18] for SRI-SCK key-points or a study of scale normalized functions for comparison of $SM$ across pyramid levels.